# Gaussian Process Regression for Out-of-Sample Extension

Oren Barkan, Jonathan Weill and Amir Averbuch

Tel Aviv University

## Abstract

Manifold learning methods are useful for high dimensional data analysis. Many of the existing methods produce a low dimensional representation that attempts to describe the intrinsic geometric structure of the original data. Typically, this process is computationally expensive and the produced embedding is limited to the training data. In many real life scenarios, the ability to produce embedding of unseen samples is essential. In this paper we propose a Bayesian non-parametric approach for out-of-sample extension. The method is based on Gaussian Process Regression and independent of the manifold learning algorithm. Additionally, the method naturally provides a measure for the degree of abnormality for a newly arrived data point that did not participate in the training process. We derive the mathematical connection between the proposed method and the Nystrom extension and show that the latter is a special case of the former. We present extensive experimental results that demonstrate the performance of the proposed method and compare it to other existing out-of-sample extension methods.

## 1. Introduction

Dimensionality reduction methods are widely used in the machine learning community for high dimensional data analysis. Manifold learning is a subclass of non-linear dimensionality reduction algorithms. These algorithms attempt to discover the low dimensional manifold that the data points have been sampled from [1]. Many manifold learning algorithms produce an embedding of high dimensional data points in a low dimensional space. In this space, the Euclidean distance indicates the affinity between the original data points with respect to the manifold geometric structure. Typically, the embedding is produced only for the training data points with no extension for out-of-sample points. Moreover, the process of computing the embedding usually involves expensive computational operations such as Singular Values Decomposition (SVD). As a result, the application of manifold learning algorithms to massive datasets or data which is accumulated over time becomes impractical. Therefore, the out-of-sample extension (OOSE) problem is a major concern for manifold learning algorithms and over the years many methods has been proposed to alleviate this problem [2, 3, 4 ,5 ,6 ,7, 22, 23].

In this paper, we propose a general framework for OOSE which is based on Gaussian Process Regression (GPR) [8]. The method is independent of the manifold learning algorithm and provides a measure of abnormality for a given test instance with respect to the training instances. The outline of the method is as follows: Given a training data and a manifold learning algorithm, we first apply the algorithm to the training data and compute the corresponding embeddings. Then, we learn the hyperparameters for a GPR model using the training data and the embeddings. Finally, given an unseen test instance and the trained GPR model, we produce a predictive distribution and set the embedding value to the distribution mode. Furthermore, the variance of the predictive distribution quantifies the degree of abnormality in the test instance. We analyze the mathematical connection between the proposed method and the Nystrom extension [9] and show that the latter is a special case of the former.

We evaluate the proposed method on several well-known manifold learning algorithms and various synthetic and real world datasets. We demonstrate its performance and show it manages to achieve competitive results when compared with other OOSE methods.

The rest of the paper is organized as follows: Section 2 overviews related work. In Section 3 we overview Gaussian Processes and GPR. Section 4 describes the proposed method and discusses its connection to the Nystrom extension [9]. In Section 5 we present experimental results.

## 2. Related work

OOSE for manifold learning is an active research field. Bengio et al. [2] proposes extensions for several well known manifold learning algorithms: Laplacian Eigenmaps (LE) [10], ISOMAP [11], Locally Linear Embeddings (LLE) [12] and Multidimensional Scaling (MDS) [13]. The extensions are based on the Nystrom extension [9], which has been widely used for manifold learning algorithms. In [3] the authors propose to use the Nystrom extension of eigenfunctions of the kernel, however in order to maintain numerical stability, they used only the significant eigenvalues. As a result, the method might suffer from inconsistencies with the in-sample data. Bermanis et al. [4] suggested to alleviate the aforementioned problem by introducing a method for extending functions using a coarse-to-fine hierarchy of the multiscale decomposition of a Gaussian kernel. The method has been shown to overcome some limitations of Nystrom extension. Recently, Aizenbud et al. [5] suggested an extension for a new data point which is based on local Principal Component Analysis (PCA).

Additional attempts to establish a solution for the OOSE problem have been taken. Fernandez et al. [6] proposes an extension of Laplacian Pyramids model that incorporates a modified Leave One Out Cross Validation (LOOCV), but avoids the large computational cost of the standard one. In [7], the authors proposed to extend the embedding to unseen samples by finding a rotation and scale transformations of the sample's nearest neighbors. Then, the embedding is computed by applying these transformations to the unseen samples. Yang et al. [20] introduced a manifold learning technique that enables OOSE by using regularization.

In the context of Bayesian statistics, Lawrence et al. [21] showed how Gaussian Process Latent Variable models can be generalized through back-constraints (GPLVMBC) to preserve local geometries. However, GPLVMBC is not designed to extend a given mapping, but to produce a new one, which is different from the original one. Moreover, the GPLVMBC requires specific derivation per objective. This is in contrast to our proposed method which learns the original mapping and hence it is independent of the manifold learning algorithm.

Wilson et al. [14] introduces a new kernel that can be used with Gaussian Processes in order to discover patterns to enable extrapolation. The new kernel was found to outperform other existing kernels. Contrary to [14], in this paper we stick to the traditional squared exponential covariance function that sometimes referred as Radial Basis Function (RBF) kernel. In our experiments, we did not observe any significant improvement when other kernels are used.

## 3. Gaussian Process Regression

Given a training set

$$D = \{(\mathbf{x}_i, y_i) \mid \mathbf{x}_i \in \mathbb{R}^N, y_i \in \mathbb{R}, i = 1,...,m\},$$

which consists of pairs of input vector and noisy predictions $(\mathbf{x}_i, y_i)$, Bayesian regression deals with computing a predictive distribution of $y_*$ for a new test instance $\mathbf{x}_*$. Typically, the noise is assumed to be additive, independent and Gaussian such that the relation between the input to the output is given by

$$y_i = f(\mathbf{x}_i) + \varepsilon_i, \quad \varepsilon_i \sim N(0, \sigma^2), \qquad (1)$$

where $f$ is a function that comes to model the noise free relation between $\mathbf{x}_i$ and $y_i$ and $N(a,b)$ stands for the normal distribution with a mean $a$ and a variance $b$.

A Gaussian Process (GP) is a stochastic process such that any finite subcollection of random variables has a multivariate Gaussian distribution. Gaussian Process Regression (GPR) is a non-parametric Bayesian regression model that assumes prior distribution of the function values such that $p(\mathbf{f} \mid \mathbf{x}_{1:m}) = N(\mathbf{0}, K_{\mathbf{ff}})$ where $\mathbf{f} = [f_1,...,f_m]^T$ ($f_i = f(\mathbf{x}_i)$) is a vector whose entries are the function values. Note that these function values are treated as random variables. $K_{\mathbf{ff}} \in \mathbb{R}^{m \times m}$ is a covariance matrix whose entries are computed by the covariance function $\left[K_{\mathbf{ff}}\right]_{ij} = \mathrm{cov}(f_i, f_j) = k(\mathbf{x}_i, \mathbf{x}_j)$. Then, for a given test vector $\mathbf{x}_*$ the predictive distribution of $y_*$ can be computed by marginalizing out the function values $\mathbf{f}$

$$p(f_* \mid \mathbf{y}) = \int p(f_*, \mathbf{f} \mid \mathbf{y}) d\mathbf{f} = p(\mathbf{y})^{-1} \int p(\mathbf{y} \mid \mathbf{f}) p(f_*, \mathbf{f}) d\mathbf{f}, \quad (2)$$

where the last transition in Eq.(2) follows Bayes rule and the fact that $\mathbf{y}$ is conditionally independent of $f_*$ given $\mathbf{f}$. Since both factors in the last integral in Eq.(2) have the following Gaussian distributions

$$p(f_*, \mathbf{f}) = N\left(\mathbf{0}, \begin{bmatrix} K_{\mathbf{ff}} & K_{f_*\mathbf{f}} \\ K_{\mathbf{f}f_*} & K_{f_*f_*} \end{bmatrix}\right), \quad p(\mathbf{y} \mid \mathbf{f}) = N(\mathbf{f}, \sigma^2 \mathbf{I}),$$

a closed form expression [8] exists for the predictive distribution

$$p(f_* \mid \mathbf{y}) = N\left(\mu_*, \sigma_*^2\right), \quad \mu_* = K_{f_*\mathbf{f}} \mathbf{A} \mathbf{y},$$

$$\sigma_*^2 = K_{f_*f_*} - K_{f_*\mathbf{f}} \mathbf{A} K_{\mathbf{f}f_*}, \quad \mathbf{A} = \left(K_{\mathbf{ff}} + \sigma^2 \mathbf{I}\right)^{-1}. \quad (3)$$

Therefore, training a GPR model amounts to the computation of $\mathbf{A}$ and $\mathbf{Ay}$. The computational complexity of the training procedure is dominated by a matrix inversion which is $O(n^3)$. Then, the prediction for a new test instance $\mathbf{x}_*$ is given by the mode of $p(f_* | \mathbf{y})$, which is the mean $\mu_*$ in case of Gaussian distribution. The variance $\sigma_*^2$ serves as a measure of the prediction uncertainty.

## 4. GPR based OOSE

Given a manifold learning algorithm $M$ and a training set $\mathbf{X} = \{\mathbf{x}_i\}_{i=1}^m \subset \mathbb{R}^N$, we apply $M$ to $\mathbf{X}$ and compute the corresponding low dimensional embedding $\mathbf{Y} = \{\mathbf{y}_i\}_{i=1}^m \subset \mathbb{R}^d (d \ll n)$. Then, for each dimension $1 \leq j \leq d$, independently, we form a new training set $D_j = \{(\mathbf{x}_i, y_{ij}) | \mathbf{x}_i \in \mathbb{R}^N, y_{ij} \in \mathbb{R}, i = 1,...,m\}$ and train a separate GPR model. Then, given an unseen test example $\mathbf{x}_*$, we predict by Eq. (3) its embedding and the measure of uncertainty in the predictions by $\mathbf{y}_* = \boldsymbol{\mu}_* = [\mu_{*1},...,\mu_{*d}]^T$ and $\boldsymbol{\sigma}_*^2 = [\sigma_{*1}^2,...,\sigma_{*d}^2]^T$, respectively. As the variance increases, our confidence in the prediction decreases and $\mathbf{x}_*$ might be considered as anomaly with respect to training set $\mathbf{X}$.

In this work we use the squared exponential covariance function (kernel)

$$\mathrm{cov}(f_i, f_j) = k(\mathbf{x}_i, \mathbf{x}_j) = \exp(-\tau^{-2} \|\mathbf{x}_i - \mathbf{x}_j\|_2^2),$$

where $\tau$ is a hyperparameter that determines the width of the kernel. In our experiments, we evaluated several other kernels and they did not produce any significant improvement. An additional hyperparameter is the noise variance $\sigma^2$ in Eq. (1). The hyperparameters can be optimized with respect to $D_j$ (note that the optimization is done for each GPR model $j$, separately). One option is to compute type II Maximum Likelihood (ML) estimates for the hyperparameters with respect to $D_j$. In the literature, this method is named as marginal likelihood [8]. Another approach is to apply cross validation. Fortunately, a close form expressions for LOOCV and its gradients exist [8] and the hyperparameters can be optimized with the Conjugate Gradient method. In this work, we use LOOCV for hyperparameter optimization. The main reason we chose this approach is that the marginal likelihood method is more prone to overfitting [8]. The algorithm is summarized in Fig.1.

---

**GPR based OOSE**

**Training phase**
*Input:*
$M$ - *manifold learning algorithm*
$\mathbf{X} = \{\mathbf{x}_i\}_{i=1}^m$ - *training set*
$d$ - *target dimensionality*
$K$ - *kernel function*
*Output:*
$G = \{G_i\}_{i=1}^d$ - *set of trained GPR models for each target dimension.*

**1.** Compute the embedding $\mathbf{Y} = \{\mathbf{y}_i\}_{i=1}^m$ using $M$ and $\mathbf{X}$.
**2.** For $j \leftarrow 1$ to $d$
  **2.1.** $D_j \leftarrow \{(\mathbf{x}_i, y_{ij}) | i = 1,...,m\}$
  **2.2.** Update $K^{(j)}$ and $\sigma^2$ (using LOOCV [8]).
  **2.3.** $\mathbf{v} \leftarrow [y_{1j},..., y_{mj}]^T$
  **2.4.** $\mathbf{A}_j \leftarrow \left(K_{\mathbf{ff}}^{(j)} + \sigma_j^2 \mathbf{I}\right)^{-1}$ (Eq. (3))
  **2.5.** $\mathbf{w}_j \leftarrow \mathbf{A}_j \mathbf{v}$
  **2.6.** $G_j \leftarrow \{\mathbf{A}_j, \mathbf{w}_j, K^{(j)}\}$

**Test phase**
*Input:*
$\mathbf{x}_* \notin \mathbf{X}$ - *test instance.*
$G = \{G_i\}_{i=1}^d$ - *set of trained GPR models for each target dimension.*
*Output:*
$\mathbf{y}_*$ - *the prediction for $\mathbf{x}_*$*
$\boldsymbol{\sigma}_*$ - *measure of uncertainty for $\mathbf{y}_*$'s entries.*

**1.** For $j \leftarrow 1$ to $d$
  **1.1.** $\mathbf{y}_{*j} \leftarrow K_{f_*\mathbf{f}}^{(j)} \mathbf{w}_j$
  **1.2.** $\boldsymbol{\sigma}_{*j} \leftarrow K_{f_*f_*}^{(j)} - K_{f_*\mathbf{f}}^{(j)} \mathbf{A}_j K_{\mathbf{f}f_*}^{(j)}$

Figure 1: GPR based OOSE algorithm

### 4.1 The connection between GPR and the Nystrom extension

Many manifolds learning methods are cast in the same framework [2], where the computation of the embedding of the training data points is obtained by eigendecomposition of a (normalized) kernel matrix. Therefore, for a given training set $\mathbf{X} = \{\mathbf{x}_i\}_{i=1}^m \subset \mathbb{R}^N$ the kernel matrix is computed by $K_{ij} = k(\mathbf{x}_i, \mathbf{x}_j)$ (this might be followed by a subsequent normalization). Then, the eigendecomposition of $K$ is carried out to form the

following relation

$$K = Y\Lambda Y^T, \quad (4)$$

where $\Lambda$ and $Y$ are a diagonal matrix with the $n$ eigenvalues on its diagonal and their corresponding column eigenvectors, respectively. Note that $K$ is a real symmetric matrix and hence $Y^T = Y^{-1}$. Finally, the embedding of $x_i$ is obtained by the $i$-th row of $Y$. We can rewrite Eq.(4) as

$$Y_{ij} = \lambda_j^{-1} K_{iX} y_j = \lambda_j^{-1} \sum_{z=1}^{n} k(x_i, x_z) Y_{zj}, \quad (5)$$

where $y_j$ is the $j$-th column eigenvector in $Y$ and $K_{iX}$ is the $i$-th row in $K$. In other words, the embedding for each data point in the training set is determined by a linear combination of the embeddings of all the other training data points multiplied by the inverse of the corresponding eigenvalue. The linear coefficients are the scaled kernel values. For the sake of simplicity, we limit the discussion to a single dimensional embedding, the generalization for multidimensional embedding is straightforward.

The Nystrom extension proposes to compute the embedding $y_*$ for a new test instance $x_*$ by

$$y_{*j} = \lambda_j^{-1} K_{*X} y_j = \lambda_j^{-1} \sum_{z=1}^{n} k(x_*, x_z) Y_{zj} \quad (6)$$

which amounts to the application of the kernel for each data point in the training set $X$ with respect to $x_*$, followed by a dot product with $y_j$.

Assuming a noise free GPR model with an identical kernel, the following relation holds: $y_i = K_{iX} K^{-1} y$ and the prediction in Eq.(3) reduces to

$$y_{*j} = \mu_* = K_{*X} K^{-1} y_j. \quad (7)$$

By using Eq. (4) we have

$$K^{-1} = \left(Y\Lambda Y^T\right)^{-1} = Y\Lambda^{-1}Y^T. \quad (8)$$

By combining Eqs. (7) and (8) we get

$$y_{*j} = K_{*X} Y\Lambda^{-1}Y^T y_j = K_{*X} Y\Lambda^{-1} e_j = \lambda_j^{-1} K_{*X} y_j. \quad (9)$$

where the second transition is due to the fact that $Y$'s columns are orthonormal and $e_j$ is the standard basis vector $j$. Notice that the predictions in Eqs. (6) and (9) are identical. Hence, the Nystrom extension is equivalent to a noise free GPR model with no hyperparamters optimization.

## 5. Experimental results

In this section we present experimental results that demonstrate the performance of our proposed method and compare it to other existing OOSE methods.

### 5.1 The experimental workflow

The workflow of the experiments is as follows: Given a manifold learning algorithm $M$, OOSE method $O$ and a dataset $X$, we apply $M$ to $X$ and derive corresponding embeddings $Y$. Then, we randomly divide $(X,Y)$ to training and test sets $R = (X_{train}, Y_{train})$ and $Q = (X_{test}, Y_{test})$, respectively. The division is done according to a specific portion $\rho$ ($\rho$ is the fraction of data points assigned to $R$ the rest are assigned to $Q$). Then, by using $O$, $R$ and $X_{test}$, we produce the embeddings $\tilde{Y}_{test}$. Finally, we measure the accuracy of the extension by the Root Mean Squared Error (RMSE) measure

$$RMSE(Y, \tilde{Y}) = \left(\frac{1}{n}\sum_{i=1}^{n} \|y_i - \tilde{y}_i\|^2\right)^{-1/2}.$$

We repeat the above procedure ten times (for different random divisions, $R$ and $Q$) to produce a series of RMSE scores and determine the final RMSE as the series average. Note that our evaluation is similar to the other previous OOSE works [4]-[7], except for the fact we add the parameter $\rho$ that challenges the evaluated methods with variable training set sizes. We will use the notations defined here throughout this section.

### 5.2 OOSE methods

We compare our proposed method to the Nystrom extension method and several OOSE methods that were recently developed and shown to overcome some of the limitations of the Nystrom extension. The methods are: Multiscale extension [4], Adaptive Laplacian Pyramids (ALP) [6] and PCA based OOSE (POOS) [5]. All of the methods provide an OOSE scheme which is independent of the manifold learning algorithm.

### 5.3 Manifold learning algorithms

We evaluate the performance of the OOSE methods for several well-known manifold learning algorithms: Diffusion Maps (DM) [3], ISOMAP [11], Laplacian Eigenmaps (LE) [10], Local Linear Embeddings (LLE) [12] and Multidimensional Scaling (MDS) [13]. We set the methods hyperparameters according to the evaluated dataset: For all datasets we applied ISOMAP, LE, LLE and MDS with the same nearest neighbor value $k = 8$. For the DM method, we adjust the neighborhood value according to the median of the squared Euclidean distances between the data points. For 3-dimensional datasets, the target dimensionality for the manifold learning methods was set to 2. For high dimensional datasets the target dimension was chosen separately, according to the spectral decay for each manifold learning algorithm.

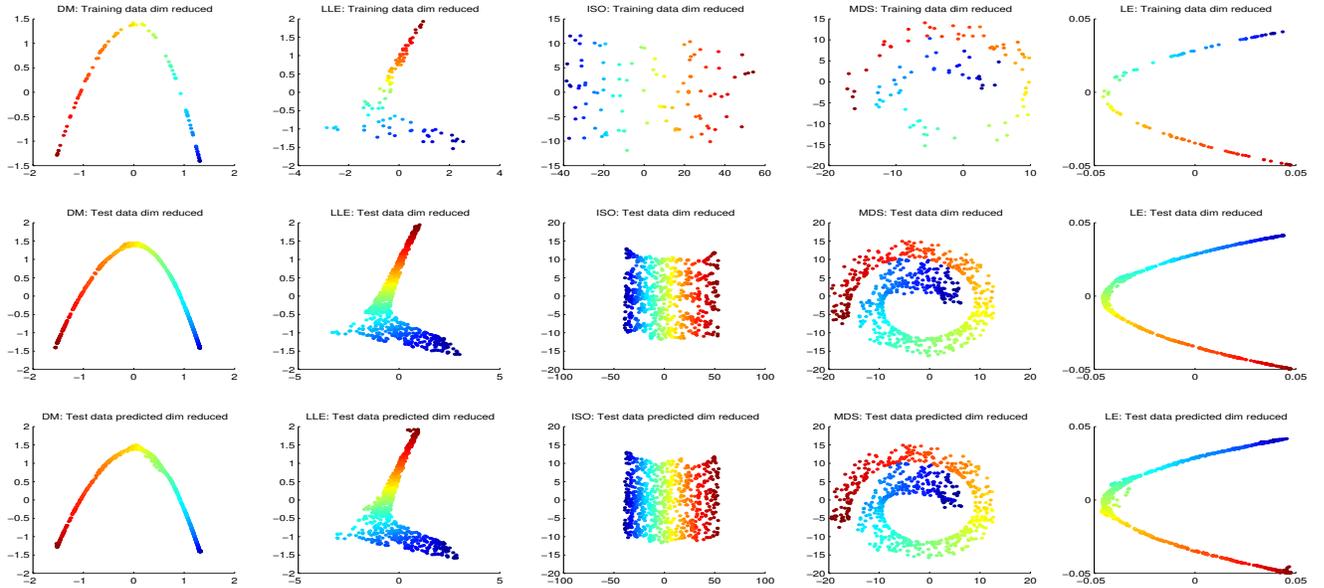

Figure 2: GPR based OOSE for various manifold learning methods and datasets. First row presents the training sets that were fed to our OOSE method. Middle row presents the true embedding that was produced by the manifold learning methods. Last row presents the embeddings produced by our method. See Section 5.5 for further details.

### 5.4 Datasets

We use various synthetic and real world datasets: Swiss roll [11], Swiss hole [15], Corner planes [16], Punctured sphere [12], Twin peaks [12], 3D Clusters [16], Toroidal Helix [3], Faces [11], MNIST [17] and USPS [18]. Each dataset poses a different challenge for the manifold learning algorithm. All datasets were fed into the manifold learning algorithms and OOSE schemes as are, without any further preprocessing. From the MNIST and USPS datasets we randomly drew a collection of 4000 and 1000 images, respectively of the same digit. For each synthetic dataset we generated a set of 1000 data points.

### 5.5 Experiment 1

Our first experiment is designed to visualize the error obtained by a GPR based OOSE. To this end, we created a Swiss roll [11] with 1000 data points (Fig.(4) left bottom corner) and produced a corresponding embedding using each of the manifold learning algorithms, separately. Then, we created $R$ and $Q$ sets with $\rho = 0.1$ as described in Section 5.1 and trained a GPR model for each of the manifold learning algorithms using $R$. Figure 2 presents the true embedding $Y_{test}$ and the predictions $\tilde{Y}_{test}$ that were produced for $X_{test}$ using GPR. As we can see, even for a small value of $\rho$, the predictions managed to preserve a small amount of noise and follow the same structure of the true embeddings $Y_{test}$.

### 5.6 Experiment 2

This experiment is designed to evaluate the OOSE methods, each time on a specific pair of a manifold learning algorithm and a dataset. To this end, given a pair $(M, X)$ we generate $R$ and $Q$ and compute average RMSE values for each $O$ (see Section 5.1 for notations and further explanation). We repeat the experiment for increasing values of $\rho$ starting from 0.05 to 0.8. Then for each $O$ we plot a graph of the log RMSE as a function of $\rho$. We used the parameters that were specified in Section 5.3. A Gaussian noise was added to all of the synthetic datasets.

The results are presented in Fig. 3. (we did not add labels to the axis, since y values are measured relatively to the competitor methods, rather than their absolute values and axis x is $\rho$ value which is clear from the context). Figure 3 is a graph table in which the $(i, j)$ entry corresponds to a specific pair $(M, X)$. The pairs are clear from the row and columns labels. As we can see GPR produces the lowest RMSE values for most of the configurations followed by POOS as the second best method. The ALP seems to perform the worse, we conclude that it is due overfitting (in the ALP algorithm, a parameter is learnt from the training data and then used in test phase [6]). The reader might notice that some of the RMSE graphs are increasing in certain late intervals (mainly for real datasets), this might be explained by outliers or instability of the manifold learning algorithm: sometimes few points in the embedding are disconnected from the rest. Therefore, as $\rho$ increases, the probability of these points to be included in $R$ increases as well.

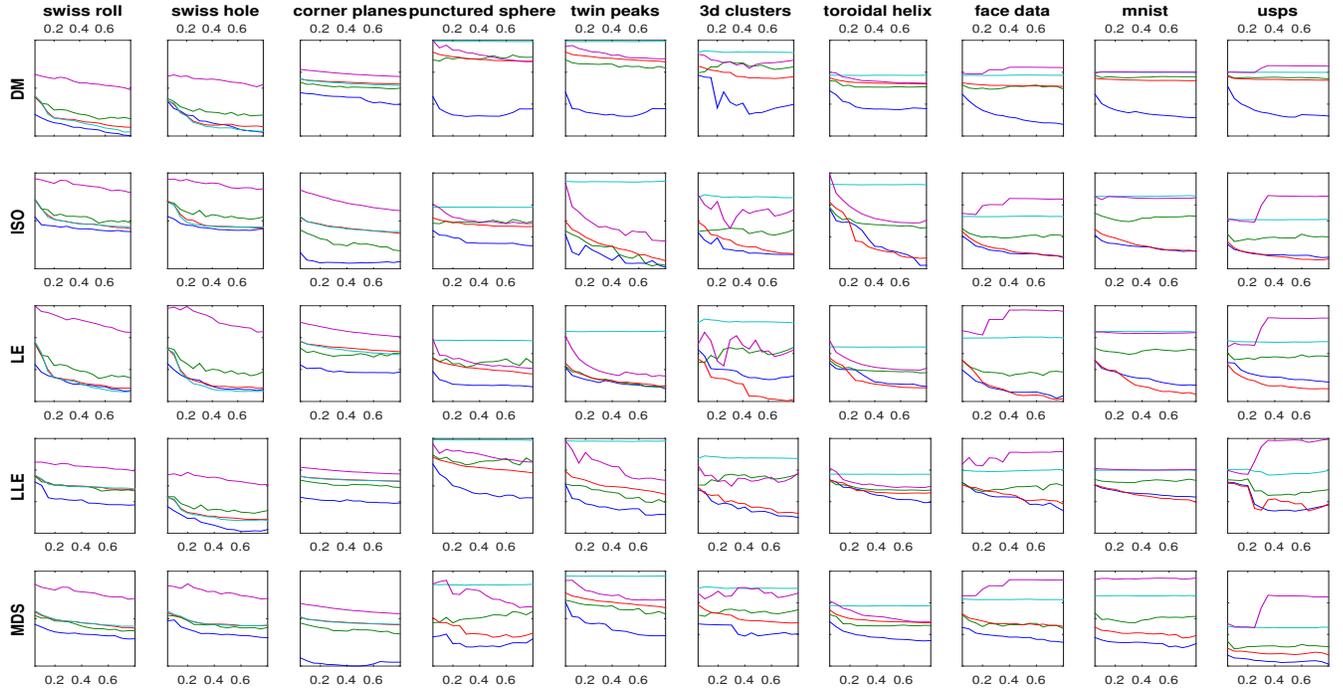
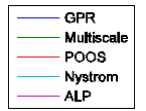

Figure 3: Table of plots of the log(RMSE) as a function of $\rho$ (the fraction of data used for training). Every plot was generated for a different combination of dataset and manifold learning algorithm. See Section 5.6 for details.

### 5.7 Experiment 3

As explained in Section 4 the GPR model produces a distribution of the prediction, with high variance $\sigma_*^2$ implies that $\mathbf{x}_*$ is anomaly. In this experiment we evaluate the GPR model as anomaly detector on synthetic datasets. We trained a GPR models for the Swiss Roll and Toroidial Helix datasets that were produced by the Diffusion Maps method (note we repeat the same experiment for the other manifold learning methods and got the same result). Then, we preserved the 2D view of the first two principal dimensions (by fixing the third dimension) and bounded it by a rectangle to form a test set, for each dataset, respectively. Figures 4 top right and bottom right show heatmaps that were produced using $H(\boldsymbol{\sigma}_*^2) = -\sum_{i=1}^{d} \sigma_{*i}^2$ for the Toroidial Helix and Swiss Roll, respectively. As we can see, for both of the datasets, the heatmaps represent the geometric structure well and anomalous points are assigned with low $H$ values.

### 5.8 Experiment 4

We experimented with using our proposed model for an anomaly detection task on the DARPA Intrusion Detection Evaluation Data Set [19]. Each instance in this dataset has 14 features based on network traffic. Every instance is associated with standard network activity or a network attack and labeled accordingly.

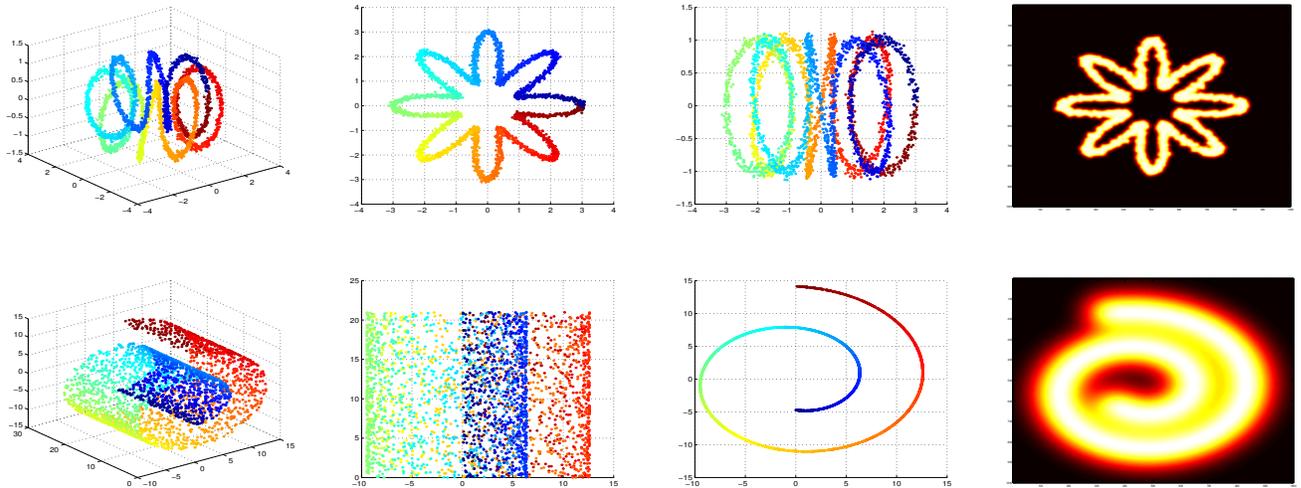

Figure 4: Plots of the toroidal helix (top) and swiss roll (bottom) views along with their corresponding heatmaps visualizing the negative variance of predictions. See Section 5.7 for details.

First, each dimension in the training set was mapped to [0, 1] using a constant scale factor. The scale factors were saved to apply the same scaling to the test instances. Next, the training data was reduced to 2 dimensions using diffusion maps with a Gaussian kernel with a neighborhood parameter $\varepsilon = 0.7$ (which was computed using the median of the average k=5 nearest neighbor distances). To decide whether a test instance is an anomaly, we use our proposed OOSE method to obtain a distribution over the 2-dimensional diffusion space of a lower dimensional vector corresponding to the test instance. The final decision is made by comparing the variance of this Gaussian distribution to a threshold. The threshold was learned by holding out 20% of the training set and optimizing the prediction accuracy on the hold out set. Using this approach, we obtained an accuracy of over 99%.

It is important to clarify that the ability to detect anomalies is a byproduct of the proposed OOSE method. We treat this capability as a **side** contribution of the paper, hence a survey of other anomaly detection methods and comparisons between them to the presented method is out of scope of this paper. This is on par with previous OOSE works such as [4]-[6]. Furthermore, though we show that our proposed OOSE method is able to achieve state-of-the-art results, we do not claim to achieve the state-of-the-art performance for anomaly detection tasks, but merely to show how to apply anomaly detection using our proposed OOSE method and validate these capabilities on both synthetic and real world datasets.

## 6. Conclusion

In this paper, we proposed a non-parametric Bayesian approach for OOSE. The method is based on GPR. We analyzed the relation between the Nystrom extension and GPR and showed that the former is a special case of the latter. We validated our proposed method in a series of experiments that demonstrated its performance and compared it to other OOSE methods. Furthermore, we showed how to apply anomaly detection using a trained GPR model and presented experimental results on both synthetic and real world datasets.

In future, we plan to investigate advanced models such as Student t processes [8] for robust Bayesian regression. We also plan to explore the performance of Relevance Vector Machines (RVMs) [24] for sparse Bayesian regression and understand whether accurate predictions can be achieved using a minimal subset of the entire training set. This might increase the computational complexity of the training phase, but substantially reduce the test runtimes. Last but not least, we plan to conduct a comparison between parametric models (e.g. neural networks) and non-parametric models for OOSE.